# A New Channel Boosted Convolutional Neural Network using Transfer Learning

Asifullah Khan[1,2*], Anabia Sohail[1], and Amna Ali[1]

[1]Pattern Recognition Lab, DCIS, PIEAS, Nilore, Islamabad 45650, Pakistan
[2]Deep Learning Lab, Center for Mathematical Sciences, PIEAS, Nilore, Islamabad 45650, Pakistan
asif@pieas.edu.pk

## Abstract

We present a novel architectural enhancement of "Channel Boosting" in a deep convolutional neural network (CNN). This idea of "Channel Boosting" exploits both the channel dimension of CNN (learning from multiple input channels) and Transfer learning (TL). TL is utilized at two different stages; channel generation and channel exploitation. In the proposed methodology, a deep CNN is boosted by various channels available through TL from already trained Deep Neural Networks, in addition to its original channel. The deep architecture of CNN then exploits the original and boosted channels down the stream for learning discriminative patterns. Churn prediction in telecom is a challenging task due to the high dimensionality and imbalanced nature of the data. Therefore, churn prediction data is used to evaluate the performance of the proposed Channel Boosted CNN (CB-CNN). In the first phase, informative discriminative features are being extracted using a stacked autoencoder, and then in the second phase, these features are combined with the original features to form Channel Boosted images. Finally, the knowledge gained by a pre-trained CNN is exploited by employing TL. The results are promising and show the ability of the "Channel Boosting" concept in learning complex classification problems by discerning even minute differences in churners and non-churners. The proposed work validates the concept observed from the evolution of recent CNN architectures that the innovative restructuring of a CNN architecture may increase the network's representative capacity.

## 1. Introduction

Deep neural networks (DNNs) are considered suitable for tackling complex learning problems. The multiple processing layers of deep architectures provide the advantage of devising multiple mapping functions for complex problems. This inherent ability has made DNNs to be one of the best tools in pattern recognition. DNNs especially, convolutional neural networks (CNNs), have shown exemplary performance on complex learning problems. CNNs are considered as effective learning models that have achieved impressive results on challenging tasks of object detection, image recognition, classification, and retrieval, etc. The high learning ability of a CNN is mainly due to two of its characteristic features; firstly, CNN learns representation in a hierarchical manner that helps to untangle complexities and learn generic representation from data. Secondly, with the increase in depth, the use of multiple mapping functions helps in handling hundreds of categories' recognition task. Moreover, high-level features can be reassigned for generic recognition tasks with no additional fine-tuning (a concept is known as multitasking). The use of a 2D convolution operator in CNN has the potential to provide effective representations of the original image. This characteristic of CNN has strengthened its ability to directly recognize the patterns from raw pixels with diminutive or no preprocessing.

The interesting work of LeCun et al. [1] brought CNNs in the limelight. Further, the widespread use of deep CNN architectures started after a classic CNN architecture AlexNet, which was proposed by Krizhevsky et al. [2]. The main contribution of AlexNet was its increased depth compared to previously proposed nets [2], which gave it improved performance over image classification tasks. This provided an intuition that with the increase in depth, the network can better approximate the target function with a number of nonlinear mappings and improved feature representation. The relationship of depth with the learning capacity gave rise to three main deep architectures, namely, VGG, Inception, and ResNet [3]. In VGG, depth is increased from 9 to 16 layers. On the other hand, in Inception networks, connections are reformulated between network layers both in terms of depth as well as in terms of the number of transformations to improve learning and representation of deep networks. Inception Nets achieved high accuracy on image data by using an interesting concept of inception modules that are comprised of multi-scale processing units (1x1, 3x3, and 5x5 convolution filters) [4]. Apart from the increase in depth of CNN, other modifications were also proposed, such as batch normalization (BN), which improved the generalization and helped reduce the diminishing gradient phenomenon in deep networks [5].

ResNet [6] introduced the interesting concept of residual learning by restructuring the architecture through identity-based skip connections, which ease the flow of information across units and, thus, helps in resolving the vanishing gradient problem. ResNet won the well-known ILSVRC competition in 2015. In 2016, Szegedy et al. (2016), combined the idea of residual connections with Inception architecture and named the proposed architecture as Inception-ResNet. They gave empirical evidence that training with skip connections can significantly accelerate the training of Inception networks [5]. Their architecture is known as Inception-ResNet.

In some networks, transformations are also defined in terms of width rather than depth. Such nets are known as Wide Residual Nets that try to solve the diminishing feature reuse problem [7]. However, in 2017, researchers started working on the improved network representation rather than the reformulation of network connections. This idea is grounded on the hypothesis that the improvement in the network representation may lead to improved classification with a low error rate [8]. Hu et al., proposed Squeeze and Excitation Nets in which the representational power of a network is improved by explicitly modeling the interdependencies between the channels (convolved feature map) [8]. In order to weigh (scale) different channels, they proposed a global averaging concept that was used to squeeze the spatial information. On the other hand, excitation function was used to excite the useful channels. In summary, recent architectures' evolution provides the concept that the restructuring of the CNN architecture to enhance representation may simplify learning with substantial improvement in performance.

In this connection, we propose a Channel Boosted deep CNN (CB-CNN) for solving complex learning problems. We hypothesize that a good representation of the input makes it easy for a modular and hierarchical learning architecture to improve its representational capability. Channel representation is boosted by augmenting the original input channel with the auxiliary learners' reconstructed output channels. Auxiliary learners are used to untangle the non-linearities among data and to learn an abstract representation of the data.  Moreover, transfer learning (TL) is exploited both for “Channel Boosting” as well as for network training. The proposed CB-CNN is evaluated on Telecom data for the prediction of churners.

Prediction of churners in telecom data is a complex classification problem due to three main reasons: (1) Telecom data has imbalanced class distribution as churners are in the minority compared to non-churners. (2) Data is high dimensional and has many ambiguous and irrelevant features such as demographic information, Call Ids, etc. (3) Unavailability of large amounts of examples to maintain the privacy of customers. The difficulty level of the churn prediction problem is also reported by several researchers who have used both traditional and deep learning techniques. Recently, DNNs have shown tremendous success in the learning of complex problems. As the churn prediction system exploits the user behavioral patterns, DNNs' application in this area can be useful not only in terms of accurate predictions but also in avoiding manual feature engineering.

The following contributions are made in this work:

- A new Channel Boosted based idea is proposed for improving representational capacity of a CNN.
- The proposed technique exploits both the dimensionality of the input channels and Transfer Learning.
- Input representation is boosted by generating diverse channels obtained through deep Generative Learners and Transfer Learning.
- The ability of Channel Boosting is demonstrated in learning complex classification problem by discerning even minute differences between the classes.

The organization of this paper is as follows. Section 2 explains the related theory. Section 3 presents the proposed methodology, and implementation details. Section 4 provides results and discussion. Finally, conclusion is made in section 5.

## 2. Related Theory of Churn Prediction

Various interesting machine learning techniques have been suggested for customer churn prediction problems; however, there is still a need for performance improvement on this challenging task. Most of the reported techniques though, highlighted the difficulty of the churn prediction problem but did not simultaneously address imbalanced distribution, high dimensionality, and training of the model. In [9], three different statistical machine learning techniques are employed, such as Decision Trees, Logit Regression, and Artificial Neural Networks (ANNs) to predict telecom churners. The reported technique, however, did not tackle the class imbalance problem. Moreover, the performance was evaluated using a few features of the dataset; samples were collected over a very short duration and were biased towards the specific geographical region [9]. On the other hand, the Support Vector Machines (SVM) based churn prediction system was proposed in [10]. SVM is a margin-based classifier, which minimizes error by drawing an optimal hyperplane in such a way that it maximizes distances between opposite classes.

The Churn prediction problem is also addressed by implementing a nonlinear instance based classifier; kNN but didn't achieve the desired performance [11]. Similarly, some researchers used nonlinear kernel methods in SVM for churn prediction, but their technique suffered from the curse of dimensionality [12]. Huang et al. [13] proposed a hybrid approach by integrating ANN, kNN, and SVM with a feature selection technique for the prediction of churners in telecom data [13]. Similarly, feature extraction techniques are also exploited in combination with the multi-combinatorial machine learning model. Seven different data mining techniques such as Linear Classifiers, Decision Trees, Naive Bayes, Logistic Regression, Neural Networks, SVM, and Evolutionary Algorithms are applied for modeling of landline customer churn prediction problem [14].

Besides individual classifiers, ensemble-based approaches are also used for the prediction of churners. Random Forest (RF) is an ensemble-based classifier that has shown satisfactory results for the prediction of churners in the banking and publishing sectors [15]. However, RF performance is still affected by the imbalance nature of the data as it draws samples using bootstrapping. This problem is overcome by modifying RF and proposed a new Improved

Balanced Random Forest (IBRF) for an imbalanced dataset [16]. IBRF iteratively learns best features by varying the class distribution and by assigning more penalty to misclassification of the minority class instance. Although IBRF showed improved accuracy compared to RF, however, its performance is only reported on a small dataset of bank customers [16]. Rotation Forest is another variant of RF that rotates the base classifier's input data by partitioning feature space in K random subsets, followed by PCA-based transformation of each feature subsets. This transformation is instrumental in bringing diversity and increasing accuracy within the ensemble [17]. Rotation forest is a widely employed model for classification tasks but generally falls short of an effective solution for forecasting telecom churners accurately. RotBoost inherits the benefit of both the Rotation forest and AdaBoost by aggregating both algorithms in one classification model [18]. RotBoost with different transformation techniques have been used for churn prediction but showed no considerable improvement in accuracy [19].

In addition to shallow learning algorithms, deep learning techniques like stacked Autoencoder (AE), and CNN are also used for churn prediction in telecom. Zaratiegui et al. encoded the customer data into images and used CNN for classification [20]. Similarly, Wangperawonga et al. proposed a hybrid stacked autoencoder (HSAE) model by integrating AE with fishers ratio analysis to enhance the classifier's learning ability [21]. Although this model gives better results than shallow algorithms, the prediction performance still needs to be improved for real-world applications.

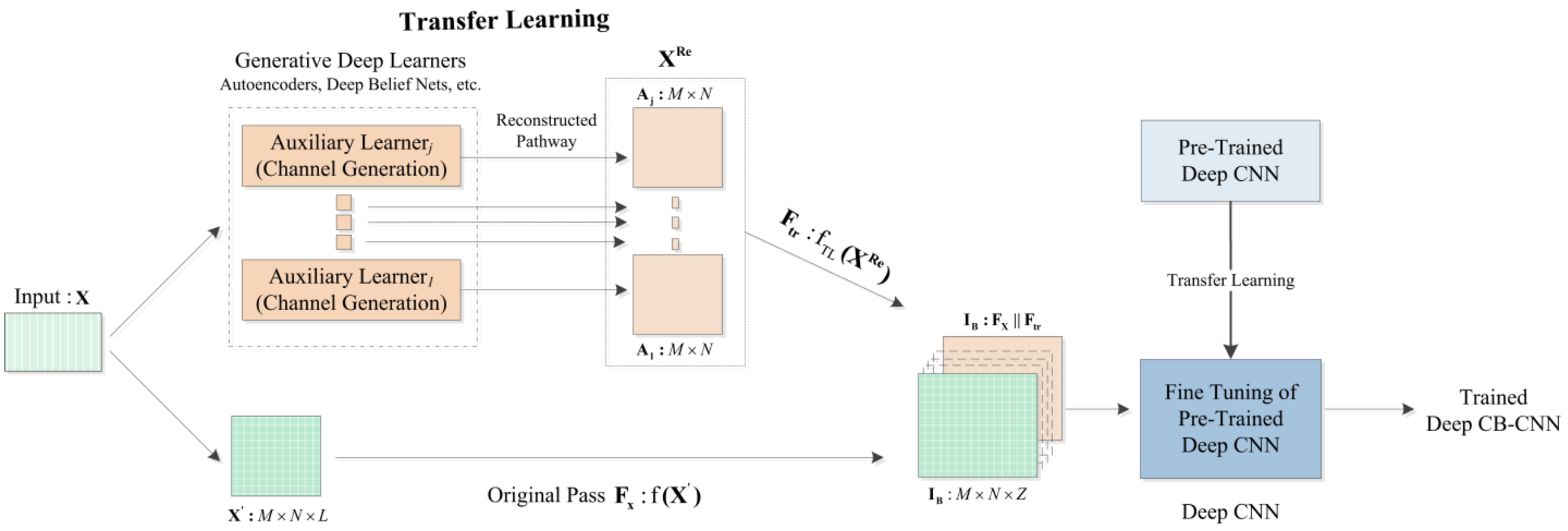

Fig. 1. Basic building block of the proposed Channel Boosted deep CNN (CB-CNN).

## 3. Proposed Methodology (CB-CNN)

In this work, we propose a novel representational enhancement known as "Channel Boosting" in deep CNN. The intuition behind "Channel Boosting" is that with the improved representation of the network through multiple input channels, we can improve the representational capacity of the CNN. The basic module of the Channel Boosted deep CNN (CB-CNN) is illustrated in Fig. 1. For any given input $\mathbf{X}$, we can construct a Channel Boosted input, $\mathbf{I_B} : \mathbf{F_x} \,||\, \mathbf{F_{tr}}$, by ensembling informative features from multiple passes. The input $\mathbf{X}$ is first passed through an auxiliary leaner, which reconstructs the input ($\mathbf{X} \rightarrow \mathbf{X^{Re}}$) in an unsupervised manner and then aggregates the original input channel $\mathbf{F_x} = \mathrm{f}(\mathbf{X})$ with the reconstructed channel $\mathbf{F_{tr}} = \mathrm{f}_{\mathrm{TL}}(\mathbf{X^{Re}})$, to produce boosted channels $\mathbf{I_B} = \mathbf{F_x} \,||\, \mathbf{F_{tr}}$. The augmentation of information (channel-wise feature representation) from multiple auxiliary learners adds the extracted information about the distribution of the data along with local and global invariance in the input representation. The auxiliary learner in a Channel Boosted module can be any generative model, whose selection mostly depends on the nature of the problem. The auxiliary learner's role is to capture complex representation from the distribution of the data and thus improve the information extraction capacity of the CB-CNN. The information generated from auxiliary learners (feature channels) is either concatenated with input feature channels or is replaced with some of the input channels with reconstructed channels to make the Channel Boosted input. In the second phase, the merits of transfer learning are utilized for the training of CNN to improve generalization and to reduce the training time. For this purpose, the pre-trained CNN, which is providing knowledge through TL, is further trained and fine-tuned with the boosted input. This further fine-tuning using the pre-trained boosted channels improves the learning ability of the network. The advantage of deploying TL is twofold: the transferred feature maps from the pre-trained CNN reduce the training cost and help in improving the generalization. On the other hand, augmentation of the auxiliary channels available through TL from the already trained Deep NN (generative model) with original feature maps improves the classifier's representation capacity for complex classification problems.

### 3.1. Implementation of the proposed CB-CNN

The performance of the proposed CB-CNN is evaluated on the telecom churn prediction problem. Telecom Churn Prediction is a complex and challenging problem owing to its imbalanced, and heterogeneous data. Details of the implementation are shown in Fig. 2.

### 3.2. Dataset and Preprocessing

Telecom data for churners and non-churners is acquired from the publicly available Orange dataset provided by the French telecom company Orange. It is a marketing database that spans over the information of the mobile network service subscribers and is used in the KDD Cup for customer relationship prediction [12]. The dataset is comprised of 50,000 samples, out of which 3672 instances are churners, while 46328 are of non-churners. Data is difficult to interpret as it is heterogeneous in nature, comprised of 230 attributes, out of which 190 are on the numeric scale, whereas 40 are categorical features. The names of the variables are masked by the organization to retain the privacy of the customers. Moreover, features are multiplied by a random factor, and categorical variables are replaced with an arbitrary value, which increased the difficulty level of the prediction.

Orange dataset constitutes heterogeneous noisy features, including variation in measurement scale (numeric to categorical variables), missing values, and sparsity of positive examples. Therefore, the data is preprocessed before the training of the classification model. Similarly, a large number of features have a significant number of missing values. Therefore, the data is divided into two subsections; features having missing values and features have no missing values. A subset of features that has more than 90% missing values are ignored, which shrinks the number of features to 208. However, for the rest of the features, the effect of missing values is compensated by replacing a missing value with the mean value of the feature column. The variation in the data is reduced by setting all the values on a positive scale. In this regard, the values on the negative scale are converted to positive scale according to (eq. 1). For this purpose, a minimum value; $m$ of a feature vector $\boldsymbol{v}_i$ is extracted. Every instance $\mathrm{x}_i$ is evaluated for its scale, if instance $\mathrm{x}_i$ lies on a negative scale, then it is converted to a positive scale, where the transformed instance is known as $y_i$.

$$m = \min(\mathbf{v_i})$$
$$y_i = x_i + m(-1) \; if \; x_i < 0 \;\text{ else }\; x_i \tag{1}$$

As the dataset comprise of both the numerical and categorical variables, therefore all the features are mapped on a numerical scale to remove disparity among the variables. Categorical features are assigned a rank by grouping them into small, medium, and large categories, based on the representation of instances in each category.

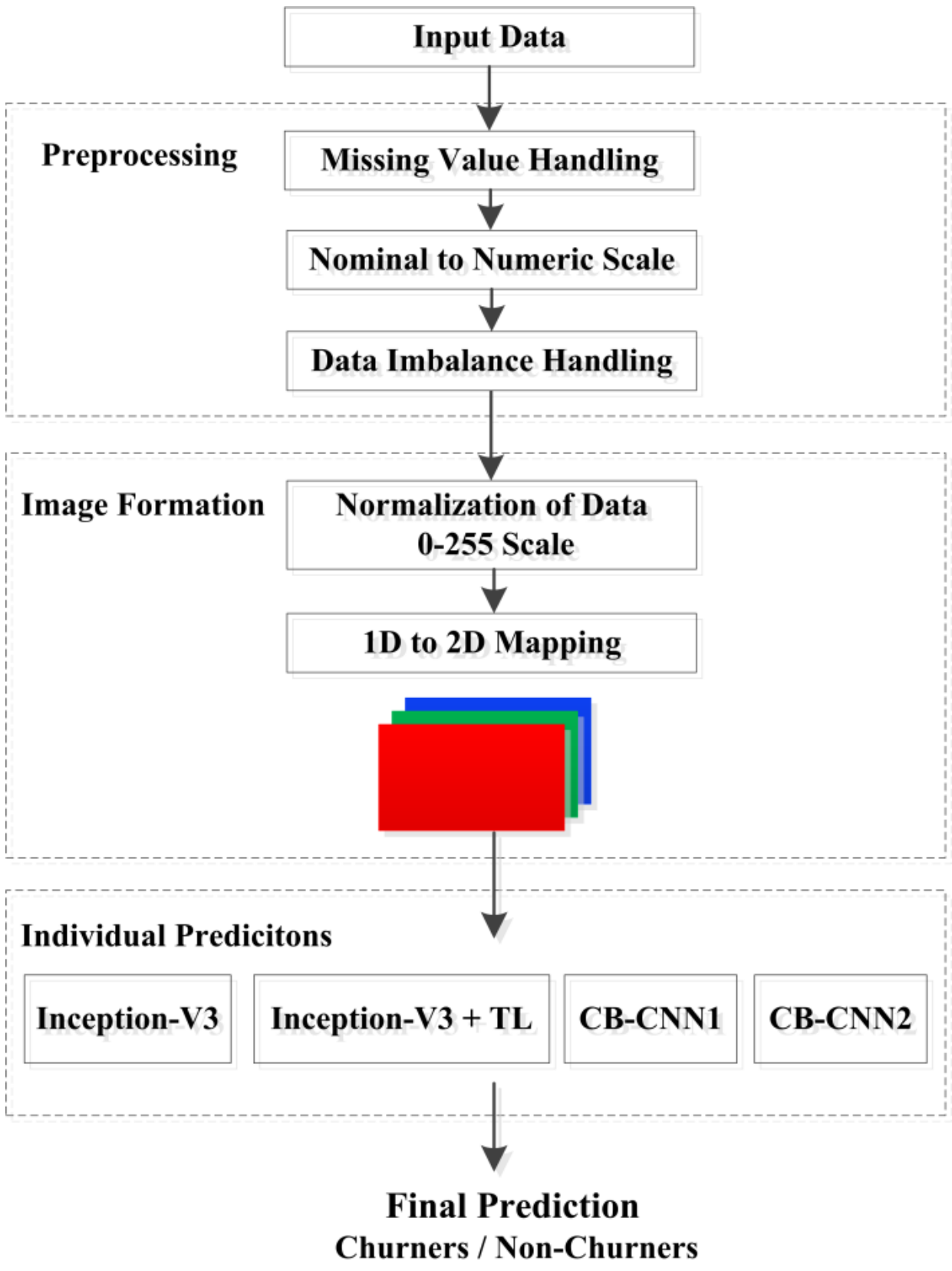

Fig. 2. Work flow of the proposed technique for churn prediction.

## 3.3. Class Imbalance

The orange dataset has a skewed distribution of churners and non-churners; it has 3672 churners (7.3%) against 46328 non-churners (92.65%), a ratio of nearly 1:13. Training a classification model on a few positive examples may affect its generalization and result in a high misclassification rate. The challenging task in highly skewed data is to detect rare events accurately. Since a rare event is usually of interest, the cost of misclassifying a positive example (churner) is high compared to that of the negative example (non-churner). The impact of class imbalance on churn prediction is reduced in the proposed technique by generating synthetic examples using Synthetic Minority Over-sampling Technique (SMOTE) [22]. SMOTE oversamples the minority class by generating synthetic examples rather than performing oversampling with replacement. SMOTE works by searching *k* number of nearest neighbors for each instance $x_i$ in the minority class [22]. One neighbor out of *k* is selected randomly, as $x_0$. Then a random number δ is generated between the range [0,1] that is used to produce a new synthetic instance, as shown in (eq. 2):

$$x_{new} = x_i + \left(x_0 - x_i\right) \times \delta \tag{2}$$

### 3.4. PCA based Dimensionality Reduction

In the proposed technique, due to high dimensionality, principal component analysis (PCA) based transformation is applied to retain the maximum variance capturing feature, while avoiding redundant correlated features. PCA based transformation reduced the 230-dimensional feature space into 60 dimensions.

PCA is a multivariate technique that is used to reduce the dimensionality of the data, while retaining maximum variance in the data. PCA transformation decomposes original dataset $(\mathbf{X} = \overrightarrow{x_1}, \overrightarrow{x_2}, ......, \overrightarrow{x_n})$ into maximum variance capturing new subspace that spans over the principal components (PC) [23].

### 3.5. Converting 1D Data into 2D Data

Orange dataset is a 1D representation of network service subscriber's information, $\mathbf{F} = \{\mathbf{f_1}, \mathbf{f_2}, ....., \mathbf{f_n}\}$. Therefore, raw information of service subscribers' is transformed into high-level feature representation by forming images (converting 1D to 2D). Image is formed by the following three steps: 1) 1D feature map is normalized to 0-255 scale using (eq. 3). 2) Rearrange the input features $\mathbf{F_i}$ to form the image such that $\mathbf{F_i} = P \times Q \times 3$, where $P \times Q$ is the size of feature channels. 3) Interpolate the intermediate pixel values and resize the image to increase the size of the image.

Before the formation of an image, the dataset is normalized using (eq. 3). For normalization, the maximum value $m_i$ of every feature vector is extracted, then every instance of $\mathbf{f_i}$ *is* divided by its maximum value $m_i$ and multiplied by 255. After normalization, every instance of the dataset is encoded to the representative image that articulates features as a pixel of the image. Initially, the size of the image is 5x5, and then images are resized, giving a final image of 32 x 32 x 3.

$$\begin{aligned} m_i &= \max(\mathbf{f_i}) \\ \mathbf{f_j} &= \left(\frac{\mathbf{f}_i}{m_i}\right)255 \end{aligned} \tag{3}$$

## 3.6. The proposed Concept of Auxiliary Learner for a CNN

In the proposed CB-CNN, generative models are used as auxiliary learners in addition to the discriminative learner (deep CNN) to boost the input representation. The use of generative models in an unsupervised way helps to characterize the input distribution (learn hidden structure) and extract useful representations. The feature representations extracted from the multiple auxiliary learners are used to build a deep hierarchy of representations (increase in width of features channels) that improves the learning ability of the machine learning model. The addition of channels in an unsupervised manner also hypothesizes that the augmented representation is invariant to small changes in the input without any preference for specific distribution of the training data, which may further improve the network performance on complex problems. The stacked AE [24] is amenable to hierarchical representation learning. Therefore, in the proposed work, 8-layers deep sparse AE (SAE) is used as an auxiliary learner to extract discriminant features. Deep SAE is trained layer by layer in an unsupervised manner using the preprocessed Orange dataset. The SAE output is used as a feature set that is converted from 1d representation to image representation.

AE is an artificial neural network that is generally used for dimensionality reduction. AE uses an unsupervised learning algorithm for modeling of input patterns [25]. It is comprised of two processing units; encoding and decoding units. The encoding unit transfers the original input $\mathbf{x}$ into input representative code $\mathbf{x_{en}}$ (eq. 4), which is then decoded by the decoding unit in such a way that there is a minimum difference between original input $\mathbf{x}$ and reconstructed input $\mathbf{d_{ec}}$ (eq. 5). The mapping learned during the encoding of original input provides the potential to learn useful features.

$$\mathbf{x_{en}} = \mathrm{f_{en}}(\mathbf{x}) \tag{4}$$

$$\mathbf{d_{ec}} = \mathrm{f_{de}}(\mathbf{x_{en}}) \tag{5}$$

The underlying architecture of AE comprises of input, hidden, and output layers. However, architecture can be deepened by stacking two or more AEs. Stacked AEs are also known as deep AE in which multiple AEs are combined, where the output of one AE is the input of the other AE. During the AE training, generally, the mean square error is used as a loss term. The intrinsic problem with traditional AE is that when it simply reconstructs the encoded input by copying information from the input layer to the output layer, it has less focus on extracting useful features. This limitation of AE can be resolved by introducing sparsity in the AE. Such type of AE is known as sparse autoencoders. Sparsity regularization forces the AE to not only perform identity mapping but also disentangle the factors of variations in the underlying data. This encourages the model to learn unique statistical features of the input, along with the encoding of input. Sparsity is induced in AE by defining a non-linear sparsity (eq.6 & 7) between linear encoder and decoder in order to improve its generalization capacity [26]. Sparsity constraint for AE is defined below:

$$\gamma_n = \gamma \tag{6}$$

$$\gamma_n = \frac{1}{k}\sum_{m=1}^{k}\left[\alpha_n(\mathbf{x}(m))\right] \tag{7}$$

where, $\gamma_n$ is the average activation of neuron $n$ whereas $\gamma$ is the sparse coefficient, whose value is set close to zero. $\alpha_n$ denotes the activation of the hidden unit $n$, when the network is given a specific input $\mathbf{x}$. Sparsity is introduced by imposing a constraint on average activation of hidden units as mentioned in (eq. 6 & 7). Constraint (eq. 6) is satisfied when hidden unit's activation value, $\gamma_n$ is close to 0. The overall cost function for SAE is given by equation (eq. 8), which is used to satisfy the constraint. KL is Kullback-Leibler divergence, whose value increases monotonically as the difference between $\gamma_n$ and $\gamma$ becomes larger. Therefore, to make information sparse, cost function penalizes $\gamma_n$ when it deviates significantly from $\gamma$ during the unsupervised learning process.

$$\Omega_{Sparsity} = \sum_{n=1}^{z} KL(\gamma \,||\, \gamma_n) \tag{8}$$

### 3.7. Deep CNN Architecture of the proposed CB-CNN

In the proposed methodology, Inception-V3 is used for the exploitation of the channel boosted input. The main strength of this architecture is due to the inception block, which convolves the same information on multiple scales in parallel by splitting the input into a multi-scale pathway and finally merges them into single concatenated output. This gives the advantage of managing information at different spatial resolutions, whereas parallel processing reduced the computational cost by introducing sparsity in the network. Inception-V3 is used to implement the proposed idea as it is reported to provide a low error rate with a reduced number of parameters and no significant loss in accuracy [27]. For the training of Inception-V3, mini-batch size is set to 20, and the learning rate of 1e-4 is used. It is run over 1000 epochs, and Adam is used as an optimizer. Adam [28] is an adaptive learning algorithm, which performs the first-order gradient-based optimization of the stochastic objective function. The objective function of Adam optimizer is given in (eq. 9). Where, $m_t$ and $v_t$ defines the first and second-order moments, respectively. The Gradient of error surface is denoted by $g_t$, whereas $\beta_1$ and $\beta_2$ are the exponentially decaying rate for the first and second-order moment estimates, respectively.

$$\begin{aligned} m_t &= \beta_1 m_{t-1} + \left(1-\beta_1\right) g_t \\ v_t &= \beta_2 v_{t-1} + \left(1-\beta_2\right) g_t^2 \end{aligned} \tag{9}$$

The details of the Inception-V3 architecture are as follows. Inception-V3 is a 159 layers deep convolutional neural network. It comprises an alternating sequence of convolution (CONV) and pooling layers and ends with a fully connected layer. All the CONV layers use a 3×3 size mask with a stride of 1 or 2. CONV layers are followed by pooling layers to reduce the spatial resolution and to provide a form of invariance to translation. Pooling is applied by using a mask of size 3x3 and 8x8 with a stride of 2. All the inputs of the CONV layers are padded with 0 to preserve the grid size. Within the inception module, three different sizes of filters (3x3, 5x5, and 1x1) are unified with a pooling layer. The output of an inception block is a concatenation of the resultant feature maps of four parallel paths. The first path consists of 1x1 convolutions, which works as a selective highway network and is used to move the selected information forward without any transformation. In the next two paths (second and third), 1x1 convolutions are accompanied by a multiscale transformation of 3x3 and 5x5 convolutions to give diverse features. In the last path, 1x1 convolution is proceeded by 3x3 pooling for the extraction of translation-invariant features. This basic module is repeated in the network, and finally, the last layer is linked to a fully connected layer. In the last layer, a fully connected layer is associated with a softmax classifier that computes the network's global loss during the training phase.

### 3.8. Transfer Learning for Channel Generation and Exploitation

In the proposed network "CB-CNN," TL is used at two stages, firstly used during the generation of the auxiliary channels through trained deep SAEs. Secondly, the painstaking process of network training is reduced by using TL for deep CNN (Inception-V3 is used as a basic learner). For channel boosting, inductive transfer learning is employed to improve the feature space. In Inductive transfer learning, both source and target domain are the same (same distribution or feature set) but are confined for different tasks. Inductive transfer learning is used to transfer the knowledge (generated channels) from auxiliary learners (source domain) to deep CNN (target domain), which then solves the learning problem through classification. This transfer of knowledge improves the feature space in the target domain, which aids in improving the learning potential of the target predictive function in the target domain.

The purpose of using TL is to boost the learning ability of a network by transferring knowledge, reducing training time, and memory consumption. Moreover, in the weight space, instead of defining random weights as a starting point, initial weights are obtained from the pre-trained network that gives a good starting point against random initialization of the weights. At first, fine-tuning of the pre-trained Inception-V3 is performed. After this, the final layer (fully connected net) of the network is trained from scratch on Orange data, as shown in Fig. 4. After training of the fully connected net, the whole network is fine-tuned using Orange dataset. This gives an advantage to the network in terms of learning local knowledge from the pre-trained net and specified knowledge of the current problem (churn prediction) using fine-tuning of the model on the boosted channels.

The concept of TL is generally applied to exploit the knowledge of an already existing model for solving new problems. In simple words, we have two domains; source domain from which knowledge is extracted and target domain, which comprises of the target problem. In TL, the reuse of knowledge is not limited by the similarity between the source and target domain; rather, this concept can be materialized if the source and target are even different but share some common characteristics [29]. Let Domain $D$ be expressed as a feature space $S$ of learning instance $E$ and a marginal probability distribution $P(E)$, where $E = \{s_1, s_2, ....s_n\} \in S$ consists of $n$ dimensional feature space. So for a given domain $D = (S, P(E))$, the task $T = \{L, \mathrm{f}(.)\}$ constitutes of label space $L$ and objective function $\mathrm{f}(.)$. The objective function $\mathrm{f}(.)$ is used to assign a label to instance $e_i$ based on conditional probability, $P(l_j \mid e_i)$, where $e_i \in E$ and $l_j \in L$. So, the source domain can be represented as $D_s = \{(e_{S_1}, l_{S_1})......(e_{S_n}, l_{S_n})\}$, whereas the target domain can be expressed as $D_T = \{(e_{T_1}, l_{T_1})......(e_{T_n}, l_{T_n})\}$. When target domain $D_T$ and source domain $D_s$ are the same, then the relation between them can be expressed as $D_T = D_s$, which implies that feature space of source and target are the same, $S_S = S_T$ or their feature space shares the same marginal probability distribution i.e. $P(E_S) = P(E_T)$. However, when $D_T \neq D_s$, it then postulates that $S_S \neq S_T$ or $P(E_S) \neq P(E_T)$ [30].

### 3.9. Implementation Details

All the experiments were carried out on the GPU enabled system with Intel processor and NVIDIA GPU: GeForce GTX 1070. Simulations were run on Python 3.5, and Matlab 2016. Autoencoders are implemented in Matlab 2016, whereas, python based deep learning library; Tensor flow and Pandas data structure are used to develop the script.

## 4. Experimental Results and Discussion

The learning potential of the proposed CB-CNN architecture is evaluated by observing its performance on the Orange dataset with and without transfer leaning. In order to exploit the learning capabilities of the 2D convolution operators in a deep CNN on telecom data, 1D customer's data is mapped to 2D images. The intensity of each pixel in the image is proportional to the customer's characteristics. Initially, the Inception-V3 network is trained from scratch on the Orange dataset. The weights are initialized randomly, and parameters are optimized using the Adam optimization algorithm. Parameters are evaluated on 10% validation data.

The generalization of the trained architecture is evaluated through 5-fold cross-validation on the unseen test dataset, which is kept separated from training data before the model optimization. Deep CNN architecture; Inception-V3 gives an AUC of 0.70 on the original feature set, without the support of any auxiliary learner. The architecture of Inception-V3 was modified to improve the final detection by reducing the false-positive rate. The depth of Inception-V3 was increased by adding more inception modules to make it deeper. It was observed that deep architecture showed improved learning for the detection of high-level features. An improvement in the precision of the model was observed as the network was made deeper, as shown in Table 1. ROC curves for Inception-V3 with 3 and 4 additional modules are shown in Fig. 3.

Next, we exploited transfer learning for the training of deep architecture. TL is used not only to improve the generalization ability but also to reduce the computation cost and training time. By exploiting TL, local information extracted from the pre-trained Inception-V3 architecture is augmented with original feature information. Churn image data is used as an input to the pre-trained Inception network, and abstraction values are computed after processing input images. These abstraction values are used to train the two fully connected layers. From the ROC curve, we conclude that Inception-V3 with TL performs better than Inception-V3 (without TL), providing an AUC of 0.73 (Fig. 3) compared to that of 0.70.

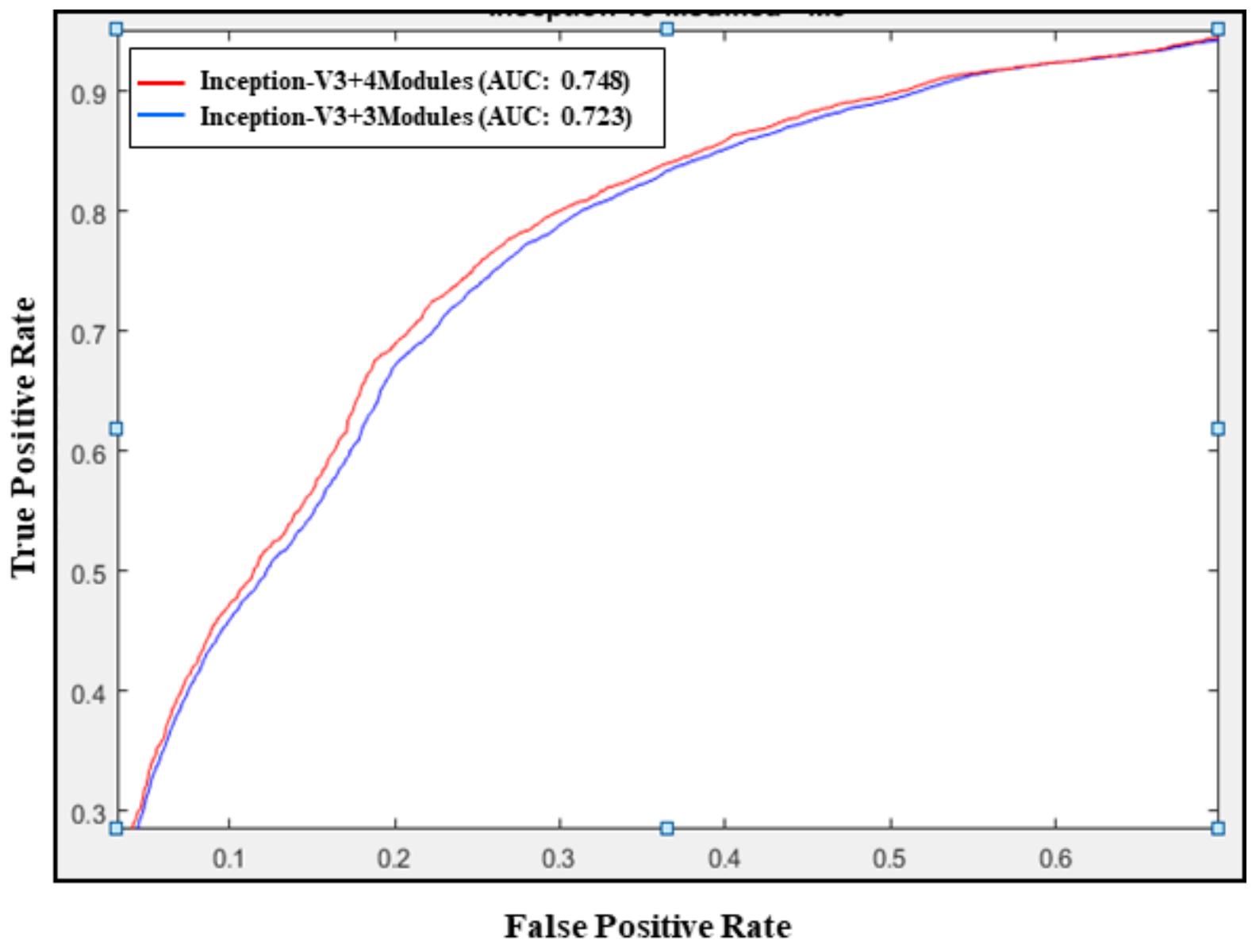

Fig. 3. ROC curve of Inception-V3 with 3 and 4 additional modules.

### 4.1. Channel Boosting and Exploitation of the Transfer Learning

The proposed CB-CNN is based on a novel architectural enhancement of "Channel Boosting" in deep CNN. This idea of "Channel Boosting" exploits the channel dimensions of CNN (learning from multiple channels) and TL's concept. The original feature representation is boosted by concatenating the original information with the reconstructed one (provided by the auxiliary learners, as shown in Fig. 4).

The reconstructed output of auxiliary learner is transformed into 2d data (images). Channel representation is boosted by replacing the channels of the original feature set with auxiliary channels. This increment in representation, thus improves the original feature representation. The pre-trained architecture is exploited through TL and fine-tuned by Channel Boosted images. Therefore, TL is utilized at two different stages; channel generation and channel exploitation. We call this specific architecture as "CB-CNN1" Net.

The performance of the proposed CB-CNN1 on telecom dataset shows improved accuracy with a boost in the channel wise information. The AUC observed for "CB-CNN1" is 0.76, and its ROC curve is shown in Fig. 5. In the next phase, we boost the channel representation by stacking the feature maps learned from auxiliary learners on the original channels (Fig. 4). The purpose of stacking is to emphasize useful features from multiple passes to make CNN easily adaptable to the complex data. Next, the pre-trained deep CNN Inception-V3 is fine-tuned on channel-augmented images (Fig. 4). The Channel boosted trained deep architecture is termed as "CB-CNN2". "CB-CNN2" gives an AUC of 0.81 and 84% accuracy, which shows that increase in channel dimensionality offers good discrimination compared to previous models trained on the original feature set.

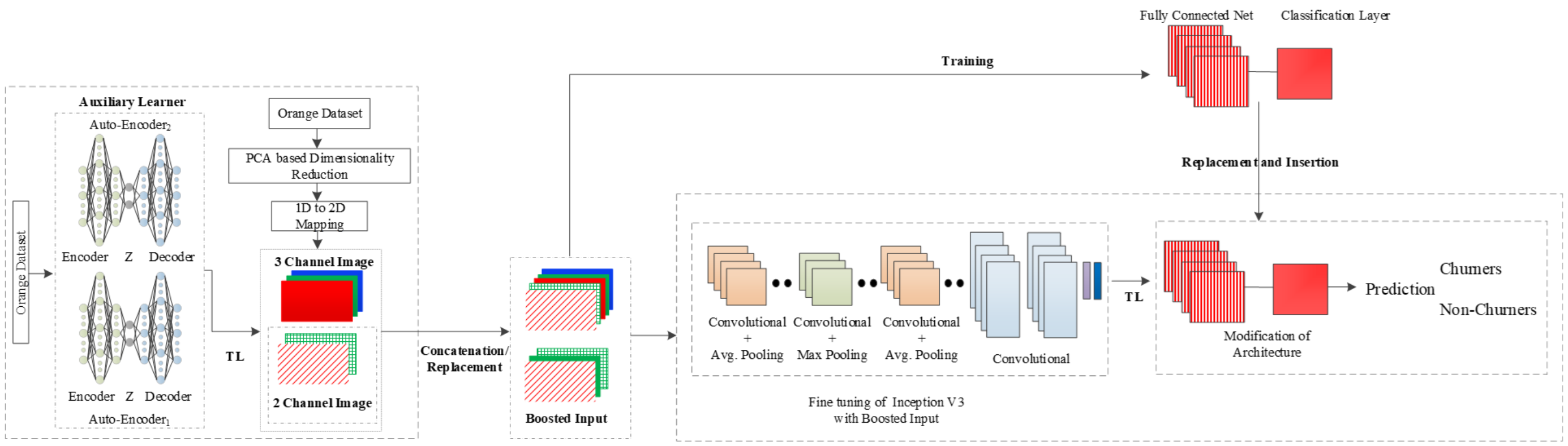

Fig. 4. Details of the working of the proposed CB-CNN. *CB-CNN1* is trained by using Boosted Input$_1$. Boosted Input$_1$ is comprised of three channels that is generated by replacing the input channels of original feature space with two auxiliary channels. Whereas, Boosted Input$_2$ is used to train *CB-CNN2* that is generated by concatenating original channel space with three auxiliary channels.

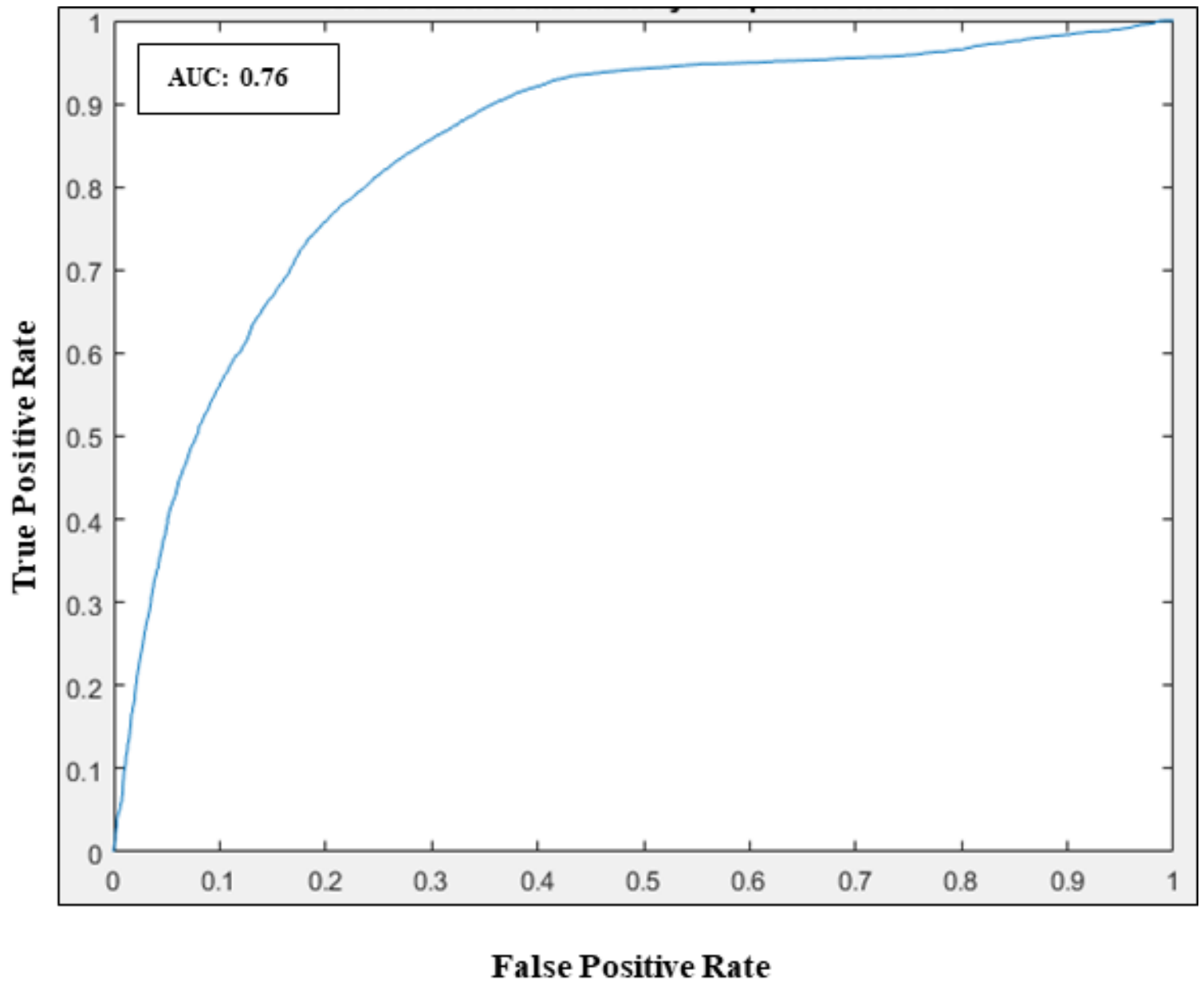

Fig. 5. ROC curve for CB-CNN1.

### 4.2. Comparison of the Proposed CB-CNN with Existing Methods

The performance of the proposed Channel Boosted architecture is compared with previously reported techniques for the Orange telecom dataset (Table 1). Both shallow and deep architecture are used for churn prediction. However, the performance of the proposed "CB-CNN2" architecture surpasses all the previously reported techniques. Idris et al. used mRMR for feature selection and exploited RF for the development of a churn prediction system [31] and reported an AUC of 0.74. In another of their work [32], they proposed a majority voting based ensemble classifier for prediction of churners. At first, they extracted the discriminant features by using filter and wrapper-based feature selection techniques. After the selection of features, an ensemble of RF, rotation forest, RotBoost, and SVM were used for prediction and reported an AUC of 0.78 using 5 fold cross-validation measure [32].

Similarly, the AUC reported by the techniques that participated in KDD Cup ranges from 0.71-0.76. Verbekea et al. [33] used a Bayesian network and reported AUC of 0.71. On the other hand, Mizil et al., [34] explored different learning algorithms such as a boosted tree, logistic regression, and RF and constructed an ensemble classifier for churn prediction problem from a library of a large number of base learners. Miller et al., [35] combined a number of shallow decision trees based on shrinkage and gradient boosting and proposed a gradient boosting mechanism. Similarly, stochastic gradient boosting based sequentially constructed regression trees "TreeNet" are reported for solving churn prediction problem [36]. All of the aforementioned techniques achieved a maximum AUC of 0.76. On the other hand, in 2017, Li used fisher ratio and deep AE for feature extraction and discrimination between churners and non-churners. They used 70% data for training and reported an AUC of 0.69 on the 30% test data. In contrast, the improved prediction of the "CB-CNN2" shows that the aggregation of information from different channels improves prediction performance over the complex dataset. Furthermore, it shows that channel-wise exploitation of information and transfer learning can improve churn prediction results.

## 5. Conclusion

In this work, we propose a novel idea of "Channel Boosting" to improve deep neural networks' representational power. We have demonstrated that the use of Channel Boosted input can improve the performance of a deep CNN. However, the idea of channel boosting can be employed in other deep neural architectures as well. Generative models are used as auxiliary learners to boost input representation. In the proposed technique, AE is used as an auxiliary learner to disentangle the input data's underlying variance and distribution. Along with this, TL is used as a bridge to make auxiliary learner's output (generated channels) available along the original feature set to deep CNN. This helps in developing a deep hierarchy of features. TL is also exploited for the final CNN network training to achieve the additional benefit of saving computational time. Simulation results show that the proposed network outperforms the reported techniques on complex and high dimensional telecom data. This suggests that the boost in input representation by stacking information at multiple levels can provide a more informative view of the input and makes model invariant to small changes. In the future, we intend to exploit generated channels not only at input layer but also to boost the intermediate feature layers of different CNN architectures. Similarly, we intend to explore, channel boosting idea by generating wavelet (frequency domain) channels for providing information-rich features. For this, the frequency domain feature will be exploited in parallel to spatial domain features downstream along with the depth of the architecture. Moreover, we also intend to utilize different auxiliary learners both in a supervised and unsupervised manner.

**Table 1:** Performance comparison of different techniques on Orange dataset.

| References | Method | AUC | Cross-validation |
|---|---|---|---|
| **Proposed method** | ***CB-CNN2*** | **0.81** | **5-fold** |
| Proposed method | *CB-CNN1* | 0.76 | 5-fold |
| Inception-V3 | Inception-V3 | 0.701 | 5-fold |
| Inception-V3 + 1 module | Inception-V3 + 1 module | 0.702 | 5-fold |
| Inception-V3 + 2 modules | Inception-V3 + 2 modules | 0.716 | 5-fold |
| Inception-V3 + 3 modules + TL | Inception-V3 + 3 modules + TL | 0.72 | 5-fold |
| Inception-V3 + 4 modules + TL | Inception-V3 + 4 modules + TL | 0.74 | 5-fold |
| Idris et al., 2015, [31] | RF and PSO based data balancing | 0.74 | 5-fold |
| Idris et al., 2017, [29] | Filter-Wrapper and Ensemble classification | 0.78 | 5-fold |
| Niculescu-Mizil et al., [34] | Ensemble Classifier | 0.76 | Hold out |
| Miller et al., [35] | Gradient Boosting Machine | 0.74 | Hold out |
| Xie et al., [36] | TreeNet | 0.7614 | Hold out |
| Verbekea et al., [33] | Bayes Network | 0.714 | Hold out |

## Acknowledgements

We thank Higher Education Commission of Pakistan for granting funds under NRPU research program (NRPU: 3408); and Pattern Recognition lab at DCIS, PIEAS, for providing us computational facilities.